# Semiconductor Defect Pattern Classification by Self-Proliferation-and-Attention Neural Network

YuanFu Yang, Min Sun, *National Tsing Hua University*

*Abstract*—**Semiconductor manufacturing is on the cusp of a revolution – the Internet of Things (IoT). With IoT we can connect all the equipment and feed information back to the factory so that quality issues can be detected. In this situation, more and more edge devices are used in wafer inspection equipment. This edge device must have the ability to quickly detect defects. Therefore, how to develop a high-efficiency architecture for automatic defect classification to be suitable for edge devices is the primary task. In this paper, we present a novel architecture that can perform defect classification in a more efficient way. The first function is self-proliferation, using a series of linear transformations to generate more feature maps at a cheaper cost. The second function is self-attention, capturing the long-range dependencies of feature map by the channel-wise and spatial-wise attention mechanism. We named this method as self-proliferation-and-attention neural network (SP&A-Net). This method has been successfully applied to various defect pattern classification tasks. Compared with other latest methods, SP&A-Net has higher accuracy and lower computation cost in many defect inspection tasks.**

## I. INTRODUCTION

Semiconductor manufacturing is a complex and time-consuming process involving hundreds of process steps. When the variation occurs in these steps, they can cause defects on the wafer surface. Automatic defect classification (ADC) is used to identify and classify the defect pattern using scanning electron microscope and provide yield engineers with important information to help them identify the root cause of die failure.

For example (as shown in Fig. 1), the defect of "Oval" mainly occurs in the crystals of the clean chamber pipeline. Once such a defect is found, we must inspect all the machine pipelines to avoid this defect from happening again. Another example is "Ball" that often occurs on the Coater. When the photoresist stops dripping, the wafer is be rotated at high speed for coating. At this time, if a small amount of photoresist continues to drip out, it will be resulting in a wafer surface with a ball shape defect.

No matter what kind of defect, we must track down the cause of the occurrence and avoid the recurrence of defect. However, in semiconductor factories, more and more edge devices are used for wafer inspection to achieve intelligent manufacturing.

This edge device allows factory managers to automatically collect and analyze data to make better-informed decisions and optimize production. It dramatically improves outcomes of manufacturing, reducing waste, speeding production, and improving yield and the quality of goods produced. Therefore, how to develop a high-efficiency architecture for ADC to be suitable for edge devices is the primary task.

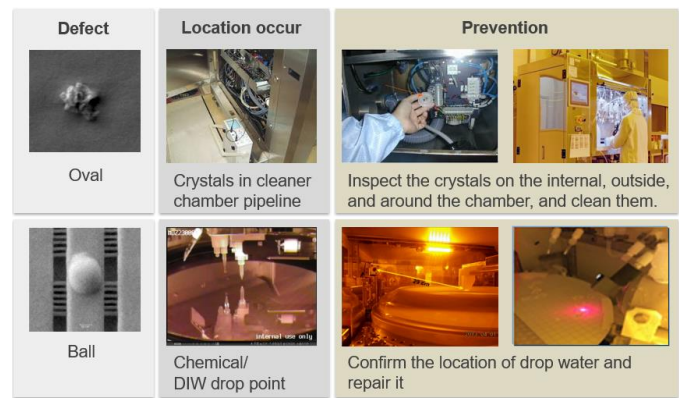

Fig. 1. Defect pattern cases and prevention procedures.

Many previous studies applied convolutional neural network (CNN) to ADC systems and became the standard approach for defect pattern classification tasks. Borisov and Scheible [1] proposed a fast and accurate solution based on novel artificial neural network (ANN) architecture for precise lithography hotspot detection using a convolutional neural network adopting a state-of-the-art technique. Nakazawa and Kulkarni [2] employed CNN for the defect pattern classification and wafer map retrieval tasks. They demonstrated that by using only synthetic data for network training, real wafer maps can be classified with high accuracy. Yang and Sun [3] propose double feature extraction method based on convolutional neural network. This model uses the Radon transform for the first feature extraction, and then input this feature into the convolutional layer for the second feature extraction for wafer map classification.

The general trend of CNNs has been to build network much deeper in order to achieve higher accuracy [4][5][6]. However, increasing the depth and complexity of the model will increase the size and computational cost, thereby making it less efficient

Min Sun is with the Electrical Engineering Department, National Tsing Hua University, Hsinchu, Taiwan, 300044 ROC (email: sunmin@ee.nthu.edu.tw).

for hardware deployment, especially in resource-constrained embedded systems. Another problem is that the feature map obtained during the convolution process is not always representative of results. This is because the spatial-wise relationship in the network has not been learned in each training (as shown in Fig. 2).

**Residual**       **Rubbing**       **Discolor**

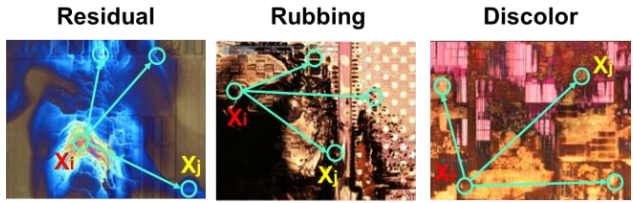

Fig. 2. A non-local operation on many defects. A position $x_i's$ response is computed by the weighted average of the features of all positions $x_j$.

In addition, the limitation of the local receptive field is also an important issue in the feature extraction [7][8]. For each convolutional layer, a set of filters are learned to express local spatial connectivity patterns along input channels. In other words, by fusing spatial and channel-wise information together, the informational combination of the convolution filter is restricted in local receptive field. Thus, CNNs can only capture the long-range dependencies with the global receptive fields by repeating a series of convolutional layers interleaved with local operation of non-linearities and down-sampling. But repeated local operations will encounter several problems. First, it is computationally inefficient. Second, it causes optimization difficulties and needs to be addressed carefully. Finally, these challenges make it difficult to model multi-hop dependencies when messages need to be delivered back and forth between distant positions [18].

In this study, we present a new network architecture that can perform feature engineering in a more efficient way. The first function is the self-proliferation, using a series of linear transformations to generate more feature maps at a cheaper cost. The second function is self-attention, learning the long-range dependencies of feature map by the channel-wise and spatial-wise attention mechanism. Compared with the latest methods, we demonstrate the success of our model (SP&A-Net) for the defect patterns classification in semiconductor wafers.

## II. RELATED WORK

In this paper, we mainly discuss the network architecture from two directions: efficiency network and self-attention. An efficient network can make full use of the computer's calculations to improve the accuracy of predictions. Self-attention mechanism can capture long-range dependencies and channel-wise dependencies to achieve gain on number of vision tasks.

### A. Efficiency Network

Since Howard et al. proposed MobileNet [9] in 2017, after generations of updates (Version-1 to Version-3), it has become an important way to learn lightweight neural networks. In MobileNetV1, they first proposed to use depth-wise separable convolution to reduce the model size (fewer parameters) and complexity (fewer multiple additions), which requires only one-eighth of the computation cost of standard convolutional neural networks.

In the MobileNetV2[10], it refers to the architecture of ResNet to add new function of expansion layer and residuals operation, which is called inverted residuals. The difference from ResNet is that MobileNetV2 first increases the dimension by pointwise convolution and then reduces the dimension after depth-wise convolution. "Extract features at high dimension and transfer information at low dimension" is the spirit of MobileNetV2. In the subsequent MobileNetV3 [11], AutoML had been applied for network architecture design. They also used linear bottleneck for inverted residuals and applied squeeze and excitation block [12] in the convolution process for feature importance learning.

GhostNet [13] first study the redundancy in feature maps and consider the redundancy in feature maps in the design of the model structure. They propose a structure, named Ghost Module, which can generate large number of feature maps with only a few calculations through a series of linear transformations. The computation cost is reduced by 10% as compared to MobileNetV3.

In this research, we refer to the architecture of MobileV3 and GhostNet to design the self-proliferation block (SP-Block). With SP-Block, we can use a cheaper operation to generate feature maps to improve the accuracy of the network.

### B. Self-Attention

The attention mechanism has been successfully applied to neural language processing in recent years [14][15]. Wang et al. [7] bridge attention mechanism and non-local operator in order to model long-range relationships in computer vision applications. Attention mechanisms can be applied along two orthogonal directions: channel-wise attention and spatial-wise attention. These two types of attention mechanisms will be introduced in the following section.

#### Channel-wise Attention

Channel-wise attention aims to model the relationships between different channels with different semantic concepts. By focusing on a part of the channels of the input feature and deactivating non-related concepts, the models can focus on the concepts of interest. Squeeze-and-Excitation Network (SENet) [12], the winner of ImageNet Large Scale Visual Recognition Challenge (ILSVRC) 2017, adaptively recalibrates channel-wise feature responses by explicitly modelling interdependencies between channels. Squeeze-and-Excitation blocks produce significant performance improvements for existing state-of-the-art deep architectures at slight computational cost. It can improve the representational capacity of a network by enabling it to perform dynamic channel-wise feature recalibration.

#### Spatial-wise Attention

As the convolutional layer builds the pixel relationship in the local neighborhood, the long-range dependences are mainly modeled by the deeply stacked convolutional layer. However, directly repeating convolutional layers is computationally inefficient and hard to optimize. This would lead to ineffective modeling of long-range dependency, due to the fact that it is difficult to deliver messages between distant positions. As the above reasons, Non-local network (NLNet) [7] proposed a pioneering method of capturing long-range dependencies, which can extract a global understanding of the visual scene. For each query location, NLNet first calculates the pairwise relationship between the query location and all locations to form an attention map, and then aggregates the features of all positions through the weighted sum with the weights defined by the attention map. Finally, the aggregated features are added to the features of each query position to form the output.

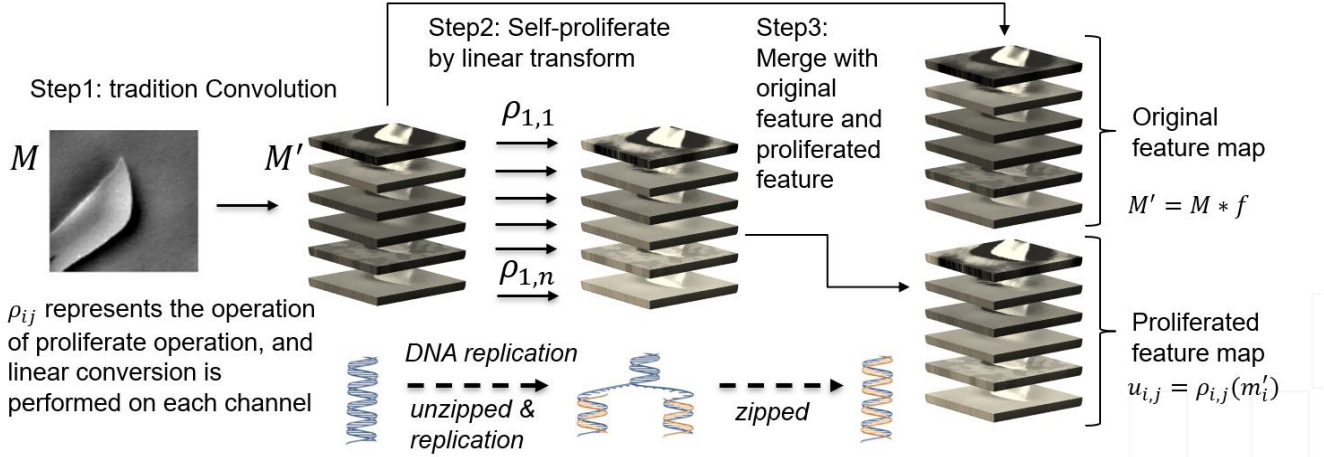

Fig. 3. Self-Proliferation Workflow.

Although Non-Local (NL) blocks have been widely used in various vision tasks [16][17], the general utilization of non-local modules has not been developed under resource-constrained scenarios such as mobile devices. This may be due to the challenge of heavy computation cost. The NL block computes the response at each position by attending on all other positions and computing the weighted average of the features in all positions, which will generate a larger computational burden. Expensive computational cost makes it difficult to apply the NL block to applications with limited computing resources.

Cao et al. [18] create a simplified network based on a query-independent formulation, which is lightweight and can effectively model the global context. This better instantiation, called the global context network (GCNet), has been shown to be superior to NLNet on major benchmarks for various recognition tasks. Li et al. [19] used neural architecture search (NAS) to design a Lightweight Non-Local (LightNL) block by squeezing the transformation operations and incorporating compact features. LightNL focus on seeking an optimal configuration of NL blocks in low-cost neural networks, which brought significant performance gains. As a result, the proposed LightNL block is usually 400 times cheaper than the conventional NL block in terms of calculation, which is favorable to be applied to mobile deep learning systems.

### III. PROPOSED APPROACH

Our network architecture is mainly composed of two parts: self-proliferation block and self-attention block. Self-proliferation block used a series of linear transformations to generate more feature maps at a cheaper cost. Then, self-attention block learned a wealth of information about long-range dependencies from this generated feature. Finally, we develop a new SP&A-Net constructed on an efficient architecture with high performance.

#### A. Self-Proliferation Block

The idea of self-proliferation operation comes from DNA replication. Proliferating cell nuclear antigen (PCNA) is an element necessary for DNA replication. PCNA achieves its processivity by encircling the DNA, where it acts as a scaffold to recruit proteins involved in DNA replication and DNA repair. The concept proposed in this paper is to increase the accuracy of the classifier by self-proliferation sufficient feature maps.

Traditional convolution is a series of convolution operations to increase the feature depth (Eq. 1.). Self-proliferation operation generates the same number of features through linear transform (Eq. 2.). The process is similar to DNA unzipping and replication, which can effectively increase the number of features (as shown in Fig. 3).

The first step of self-proliferation is a convolution operation which can be built upon a transformation $f \in \mathbb{R}^{c \times k \times k \times n}$ mapping an input $M \in \mathbb{R}^{c \times h \times w}$ to feature maps $M' \in \mathbb{R}^{h' \times w' \times n}$:

$$M' = M \otimes f, M \in \mathbb{R}^{c \times h \times w}, f \in \mathbb{R}^{c \times k \times k \times n}, M' \in \mathbb{R}^{h' \times w' \times n} \quad (1)$$

where $c$ is the number of input channels, $h$ and $w$ is the height and width of the input data. Here $\otimes$ denotes convolution.

The second step is a cheap operation, represented by $\rho$ in the Eq. 2. This is a linear transformation that uses depth-wise operations to generate the other half of the feature map:

$$u_{i,j} = \rho_{i,j}(m'_i), \forall i = 1, \dots, s, j = 1, \dots, n \quad (2)$$

where $i$ and $j$ represents the position $(i,j)$ on the feature map, $m'_i$ is the unit of feature map $M'$ generated by Eq. 1., $s$ and $n$ is the height and width of new feature map $u_{i,j}$. If 2n convolution kernels are used normally, n convolution kernels are used for self-proliferation. Therefore, the computation cost can be reduced by half (as shown in Fig. 3).

*B. Self-Attention*

The self-attention block aims at strengthening the features of the query position via aggregating information from other positions. The basic architecture is formulated as follows:

$$y_i = x_i + w_2 ReLU(LN\left(w_1 \sum_{j=1}^{h \times w} \frac{e^{w_g x_j}}{\sum_{m=1}^{h \times w} e^{w_g x_m}} x_j\right)) \quad (3)$$

where $x_i$ denotes the feature map of one input instance, e.g., defect pattern, $h$ and $w$ is the height and width of input $x$, $w_1$ and $w_2$ denote linear transform matrices ($1 \times 1$ convolution) which is used to bottleneck transform, $w_g$ is the weight for global attention pooling, $LN$ denotes the layer normalization, which is utilized to filter the redundant information and refine the obtained contextual information. Specifically, our self-attention block consists of 3 parts (as show in Fig. 4): (a) global attention pooling for context modeling; (b) bottleneck transform to capture channel-wise dependencies; and (c) denotes the fusion function to broadcast element-wise addition for feature fusion. This function can aggregate the global context features to the features of each position [18].

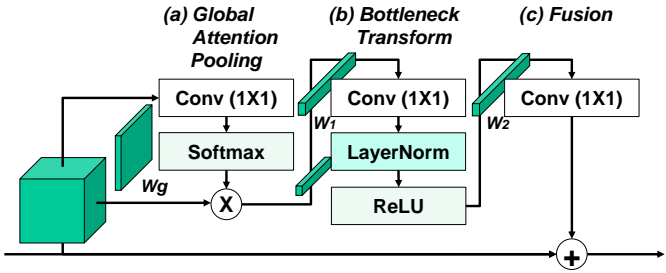

Fig. 4. Self-Attention Architecture.

*C. Self-Proliferation-and-Attention Block*

Our self-proliferation-and-attention block consists of 4 parts (as shown in Fig. 5): (a) expansion layer used self-proliferation block to increase input dimension to generate more feature maps at a cheaper cost. (b) depth-wise convolution is used for feature extraction and self-attention are used to capture long-range dependencies. (c) compression layer used to reduce feature dimension to be the same as input feature. Then, we can transform spatial information with low computational effort. In addition, we also remove the activation of ReLU in the compression layer to avoid the tensor collapse problem [10]. Finally, in (d) we refer to MobileNetV2 [10] to build inverted residuals. The residuals architecture was first proposed by He et al. in ResNet [20]. Its design concept is to learn residuals on the network to avoid the problems of gradient vanishing. ResNet's architecture is to compress first then expand. MobileNet's architecture is the opposite of ResNet that expand first then compress. The idea of MobileNet is to "capture features in high dimensions and transfer information in low dimensions".

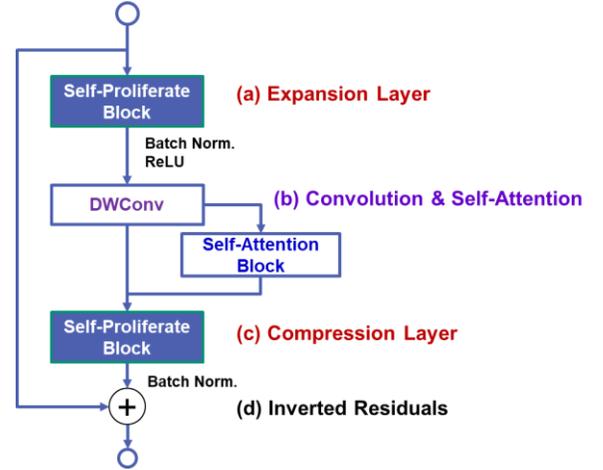

Fig. 5. Self-Proliferation-and-Attention Architecture.

*D. Self-Proliferation-and-Attention Neural Network*

In the overall network design of SP&A-Net, we use the network architecture designed by Auto ML [11] due to its superior performance. As Table I, we set a total of 16 layers of SP&A block. Self-Attention block is applied to all self-proliferation layer. At the end, Conv2D $1\times1$ is used to increase dimension to 1280. In the flattening layer, 1000 features are output to connect the classes of defect labels.

TABLE I
SP&A-NET ARCHITECTURE

| Input | Block | Output | Expansion | Self-Attention |
|---|---|---|---|---|
| 512X512X3 | Conv2D 3X3 | 16 | - | - |
| 256X256X16 | SP&A Block | 16 | 16 | 1 |
| 256X256X16 | SP&A Block | 24 | 48 | 1 |
| 128X128X24 | SP&A Block | 24 | 72 | 1 |
| 128X128X24 | SP&A Block | 48 | 72 | 1 |
| 64X64X48 | SP&A Block | 48 | 120 | 1 |
| 64X64X48 | SP&A Block | 96 | 240 | 1 |
| 32X32X96 | SP&A Block | 96 | 200 | 1 |
| 32X32X96 | SP&A Block | 96 | 184 | 1 |
| 32X32X96 | SP&A Block | 96 | 184 | 1 |
| 32X32X96 | SP&A Block | 144 | 480 | 1 |
| 32X32X144 | SP&A Block | 144 | 672 | 1 |
| 32X32X144 | SP&A Block | 192 | 672 | 1 |
| 16X16X192 | SP&A Block | 192 | 960 | 1 |
| 16X16X192 | SP&A Block | 192 | 960 | 1 |
| 16X16X192 | SP&A Block | 192 | 960 | 1 |
| 16X16X192 | SP&A Block | 192 | 960 | 1 |
| 16X16X192 | Conv2D 1X1 | 960 | - | - |
| 16X16X960 | AvgPool 16X16 | - | - | - |
| 16X16X960 | Conv2D 1X1 | 1280 | - | - |
| 1X1X1280 | FC | 1000 | - | - |

## E. Algorithm for Model Training

Our SP&A-Net used minibatch Nesterov Accelerated Gradient (NAG) descent for training. In each iteration, we randomly sample n images to calculate the gradient, then update the network parameters. It stops after k passes through the dataset. All functions and hyperparameters in our algorithm can be implemented in different neural networks introduced before. The formal algorithm can be formalized as follows.

TABLE II
ALGORITHM OF SP&A-NET

| Algorithm: Train a Self-Proliferation with mini-batch NAG descent |
| --- |
| Inputs: defect pattern image (**h x w**)<br>Require: convolution mapping **M**, out channel **O**, dimension reduce **R** = 50%, Conv1X1 kernel size **convkernel** = 1X1, strides **S** = (1,1), Linear operation **LO**, Linear operation kernel **LK** =3X3<br>Initialize(**SP&A-Net**)<br>for epoch = 1, 2, ...., k **do**<br>    for batch = 1, 2, ...., #images/b **do**<br>        ***images*** ← uniformly random sample b images<br>        **X, y** ← preprocess(***images***)<br>        **O = M * R**<br>        $I^{st}$**_conv** = conv(**O, convkernel, S**)<br>        **LO** = depthwise($I^{st}$**_conv, LK**)<br>        ***Output*** = concatenate($I^{st}$**_conv, LO**)<br>        *f* ← forward(**SP&A-Net, X**)<br>        *L* ← **Cycleloss**(*f, y*)<br>        *grad* ← backward(*L*)<br>        update(**SP&A-Net, grad**)<br>    end for<br>end for<br>Return ***Loss, Accuracy*** |

The weights of all layers are initialized with the Xavier algorithm [21]. In addition, we use cycle loss proposed by Sun et al. [22] to estimate the loss rate during model training. The circle loss has a unified formula for two elemental deep feature learning approaches, i.e. learning with class-level labels and pair-wise labels. Its goal is to maximize the within-class similarity ($s_p$) and minimize the between-class similarity ($s_n$), as shown in formula (4):

$$L_{circle} = \log[1 + \sum_{i=1}^{K}\sum_{j=1}^{L} \exp(\gamma(\alpha_n^j s_n^j - \alpha_p^i s_p^i))] \quad (4)$$

where $\alpha_n^j$ and $\alpha_p^i$ are non-negative weighting factors, $K$ is the within-class scores, and $L$ is between-class similarity scores with a single sample in the feature space. The scale factor $\gamma$ determines the largest scale of each similarity score. Circle loss on $\gamma$ is the automatic attenuation of gradients. As the similarity scores approach the optimum during training, the weighting factors gradually decrease. Consequentially, the gradients automatically decay, leading to a moderated optimization. Cycle Loss generalizes ($s_n$-$s_p$) to ($\alpha_n s_n - \alpha_p s_p$), allowing $s_n$ and $s_p$ to learn at different paces.

## IV. EXPERIMENT RESULTS

To evaluate the proposed method, we carry out experiments on defect pattern datasets on 3 types of defect inspection, including AEI (After Etch Inspection), ADI (After Development Inspection), and API (After Polish Inspection). In addition, we also verify our models in public databases, including of CIFAR-10 and ImageNet ILSVRC 2012 dataset.

### A. Data Exploration

In the defect pattern dataset, there are total of 11 defect types, include of Remain, Silk, Multi-dots, Scratch, Small-particle,

TABLE III
DATA EXPLORATION IN DEFECT PATTERN

| Defect Type | Image1 | Image2 | Image3 | Feature Description | Defect Type | Image1 | Image2 | Image3 | Feature Description |
| --- | --- | --- | --- | --- | --- | --- | --- | --- | --- |
| Remain | 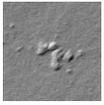 | 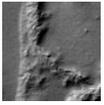 | 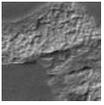 | Flat irregular shape, usually on the edge of the wafer | Hump | 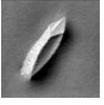 | 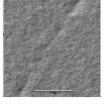 | 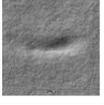 | Irregular bumps, edges fused to surface |
| Silk | 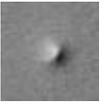 | 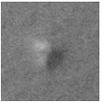 | 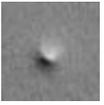 | Discontinuous linear protrusion | Flask | 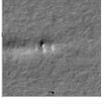 | 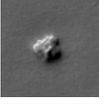 | 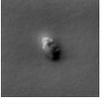 | Flaky, undulated top surface |
| Multi-dots | 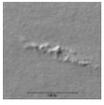 | 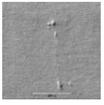 | 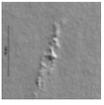 | Random dots raised (more than three points) | Fallon | 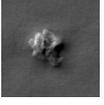 | 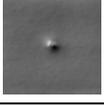 | 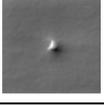 | Large irregular particles |
| Scratch | 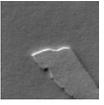 | 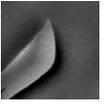 | 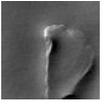 | Linear depression | Oval | 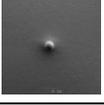 | 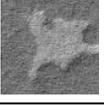 | 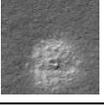 | Large particles made of spheres |
| Small-Particle | 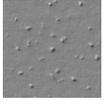 | 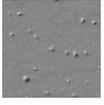 | 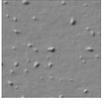 | Small irregular particles | Color Mark | 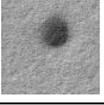 | 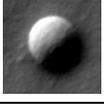 | 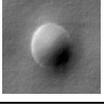 | No obvious undulations on the surface |
| Ball | 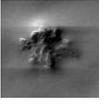 | 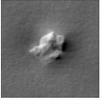 | 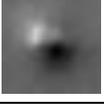 | Round particles | | | | | |

Ball, Hump, Flask, Fallon, Oval, and Color mark (as shown in Table III). Each defect has different characteristics and elements. For example, the defect type of " Flask " is composed of Aluminum (Al) and Oxygen (O), the defect type of " Falon " is composed of Ferrum (Fe) and Nickel (Ni), and the defect type of " Oval " is composed of Fluorine (F).

Aging of different machine parts will cause different defects. For example, the defect of " Flask ", which contains the elements of Aluminum (Al) and Oxygen (O), mainly comes from the electrostatic chuck (ESC) in the chamber of the Chemical Vapor Deposition (CVD) process (as shown in Fig. 6). The electrostatic chuck is a device for generating an attracting force between an electrode and an object at a voltage applied to the electrode. Under the high temperature and high-pressure environment of the CVD process, particles will fall on the wafer surface as the electrostatic chuck get aged. When we detect this kind of defect, meaning that the chamber should be maintained and aging parts placed by new ones. Similarly, when other types of defect have been detected, they must have their own out of control action plan (OCAP) to ensure the quality of semiconductor manufacturing.

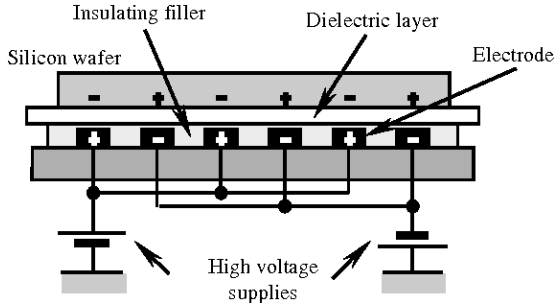

Fig. 6. The structure of electrostatic chuck (ESC)

## B. Ablation Study

We first randomly selected 10,000 images from 3 kinds of defect pattern datasets (AEI, ADI, and API) for ablation study. Then, we select 6,000 images for training, 2,000 images for validation, and 2,000 images for testing. Table IV shows the ablation study of the SP&A-Net compared with the ResNet-50 in different blocks. GC block performs slightly better than SE, NL and SNL blocks. SP&A-Net performs better than ResNet-50 with GC blocks. According to the ablation study results, we choose Layer Normalization as the regularization strategy, Conv1×1 as the flattening method, Nesterov Accelerated Gradient as the optimizer strategy, and finally Cosine Annealing as the learning rate function.

## Ablation Study on the Composition Ratio

Table V presents the accuracy, precision, recall, F1-score, and model parameters in 7 composition ratio (r) of self-proliferation in defect pattern classification. We do experiments under the same number of SP&A layers (16 layers). In order to prevent the classifier from not detecting defective wafers (false negative), the internal rule is that the recall rate should be greater than 99%. Therefore, we exclude parameter settings with r greater than 0.5. In addition, the accuracy of r = 0.06 / 0.13 / 0.25 is slightly higher than the accuracy of r = 0.05. However, the number of parameters used in its model has increased significantly, so we take r = 0.5 as the best parameter setting.

TABLE V
ABLATION STUDY ON THE COMPOSITION RATIO OF SP&A-NET

| r | Accuracy | Precision | Recall | F1-Score | #Params (M) |
|---|----------|-----------|--------|----------|-------------|
| 0.06 | 98.45% | 97.45% | 99.16% | 98.30% | 5.10 |
| 0.13 | 98.44% | 97.43% | 99.15% | 98.28% | 3.40 |
| 0.25 | 98.40% | 97.37% | 99.05% | 98.20% | 2.98 |
| **0.50** | **98.38%** | **97.32%** | **99.03%** | **98.17%** | **2.60** |
| 0.63 | 97.81% | 96.75% | 98.44% | 97.58% | 2.47 |
| 0.83 | 97.21% | 96.38% | 97.48% | 96.93% | 2.35 |
| 1.00 | 95.83% | 94.24% | 96.62% | 95.42% | 2.29 |

## C. Experimental Results on Defect Pattern Datasets

To objectively verify our proposed model, we conduct experiments on 3 defect pattern datasets, including AEI (After Etch Inspection), ADI (After Development Inspection), and

TABLE IV
ABLATION STUDY WITH SELF-ATTENTION, SELF-PROLIFERATION, AND SELF-PROLIFERATION-AND-ATTENTION NEURAL NETWORKS

| Model | #Params (K) | Refinement | | | | | | | | | |
|-------|-------------|------------|---|---|---|---|---|---|---|---|---|
| | | Regularization | | | Flatten | | Optimizer | | | Learning Rate | |
| | | Batch Normalization | Layer Normalization | Group Normalization | Global Avg. Pooling | Conv1X1 | SGD | Adam | NAG | Reduce LR On Plateau | Cosine Annealing |
| ResNet50+SE Block | 2673 | 97.23% | 97.23% | 97.10% | 97.07% | 97.25% | 97.03% | 97.23% | 97.12% | 97.31% | 97.39% |
| ResNet50+NL Block | 4125 | 97.44% | 97.41% | 97.51% | 97.28% | 97.58% | 97.44% | 97.54% | 97.59% | 97.35% | 97.38% |
| ResNet50+SNL Block | 3742 | 97.46% | 97.57% | 97.61% | 97.62% | 97.34% | 97.21% | 97.35% | 97.38% | 97.52% | 97.25% |
| ResNet50+GC Block | 3353 | 97.45% | 97.19% | 97.34% | 97.08% | 97.44% | 97.42% | 97.31% | 97.51% | 97.24% | 97.66% |
| SPNet (without attention) | 2120 | 97.58% | 97.64% | 97.59% | 97.49% | 97.67% | 97.37% | 97.39% | 97.49% | 97.54% | 97.57% |
| SP&A-Net (with self-attention) | 2135 | 97.94% | 98.15% | 97.93% | 97.91% | 97.94% | 97.66% | 97.69% | 97.87% | 97.97% | **98.27%** |

Five self-attention models: SE-ResNet (ResNet with SE Block), Non-Local-ResNet (ResNet with NL Block), Simplified-Non-Local-ResNet (ResNet with SNL Block), Global-Context-ResNet (ResNet with GCBlock), and SP&A-Net (Self-Proliferation-and-Attention-Network). SPNet: Self-Proliferation-Network, without Self-Attention.

API (After Polish Inspection). During the experiment, we compared the accuracy of the models with the number of parameters.

*AEI Datasets*

The AEI defect pattern dataset is provided by a scanning electron microscope (SEM) inspection applied after the etch process. In Fig. 7., we have designed 3 scenarios for accuracy and parameters to evaluate the performance of the following five models. From the results, we can see that generally larger parameters lead to higher accuracy in these 5 networks which shows the effectiveness of them. In addition, SP&A-Net has a higher accuracy than other model under the same number of parameters.

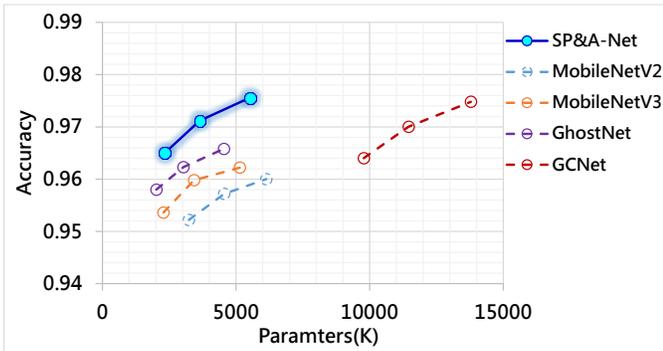

Fig. 7. Experimental result of AEI defect pattern

*ADI Datasets*

ADI complements the lithography defect reduction program by capturing topography-related process-integration defects, and it can check reticle integrity through defect identification. In ADI, we have also designed 3 scenarios for accuracy and parameters to evaluate the performance of five models. We can see that SP&A-Net outperform the baseline across all parameters (as shown in Fig. 8).

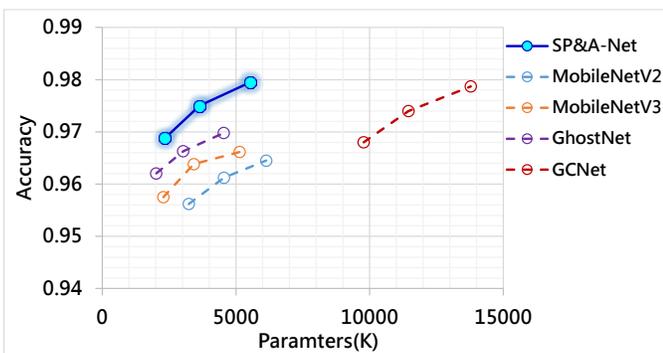

Fig. 8. Experimental result of ADI defect pattern

*API Datasets*

We further assess the generalization of SP&A-Net on the task of defect detection using the API dataset. The experiment showed the same results as before in Fig. 9. SP&A-Net has a higher accuracy than other models under the same number of parameters.

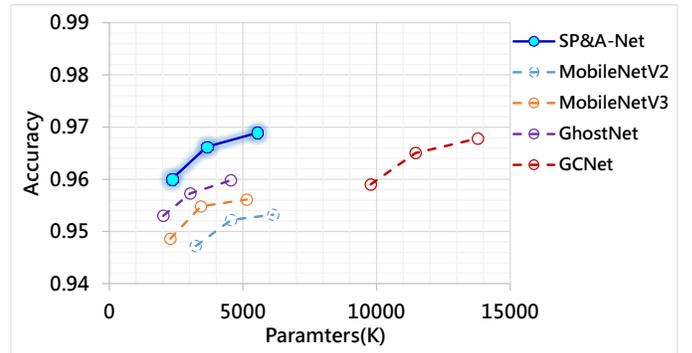

Fig. 9. Experimental result of API defect pattern

### D. Experiment Results of Each Defect Type

Table VI shows the performance of SP&A-Net for 11 defect types. We use F1-Score as an evaluation metric instead of accuracy to avoid the data unbalance issue. F1-Score of each defect types are high than 97%, except for defect type of "Silk" and "Multi-dots" . It may be due to the uncertain structure, which is easy to be confused with other defect shapes.

TABLE VI
EXPERIMENT RESULTS OF EACH DEFECT TYPE

| Defect Type | AEI | ADI | AXI |
|---|---|---|---|
| Remain | 98.5% | 98.0% | 98.4% |
| Silk | 97.0% | 96.2% | 96.6% |
| Multi-dots | 96.7% | 97.0% | 96.3% |
| Scratch | 97.5% | 98.3% | 97.7% |
| Small- | 99.0% | 98.4% | 98.1% |
| Particle | 98.2% | 98.8% | 98.9% |
| Ball | 98.8% | 98.9% | 98.8% |
| Hump | 97.3% | 97.8% | 97.6% |
| Flask | 98.7% | 98.3% | 98.7% |
| Fallon | 98.4% | 98.6% | 98.2% |
| Oval | 98.4% | 98.3% | 98.7% |
| Color | 98.3% | 98.5% | 98.7% |
| Mark | 98.1% | 98.8% | 98.2% |

### E. Experiment on the Public Datasets

This paper is mainly for model verification on the defect pattern dataset of the semiconductor factory. In addition, we are very curious about the proposed model on the real environment image. To objectively verify our proposed model, we conduct experiments on 2 public visual datasets, including CIFAR-10 and ImageNet ILSVRC 2012 dataset.

*CIFAR-10*

CIFAR-10 consists of 60,000 32×32 color images in 10 classes, with 50,000 training images and 10,000 test images. We compare SP-Net and SP&A-Net with several representative models on ResNet-50 architectures (as Table VII). Our self-proliferation model outperforms the competitors with the highest performance (92.92%/92.93) but with significantly fewer parameters. But from the results, we can see that there is no significant difference between the accuracy of SPNet and SP&A-Net. This is because the image size is too small (32×32) that the attention mechanism has no chance to show its advantages.

TABLE VII
COMPARISON OF STATE-OF-THE-ART METHODS ON CIFAR-10

| Model | Accuracy | #Params (M) | FLOPs (M) |
|---|---|---|---|
| ResNet50+SE Block | 92.10% | 0.83 | 127.0 |
| ResNet50+NL Block | 92.04% | 0.92 | 142.2 |
| ResNet50+SNL Block | 91.91% | 0.78 | 120.4 |
| ResNet50+GC Block | 92.66% | 0.82 | 126.3 |
| SPNet (without attention) | 92.92% | 0.41 | 94.8 |
| SP&A-Net (with self-attention) | 92.93% | 0.42 | 95.1 |

*ImageNet ILSVRC 2012 dataset*

ImageNet is a large image dataset containing more than 1.2 million images. In this paper, we directly perform our model on the ImageNet training set. Usually, we keep 50K images randomly selected from the training set as a fixed validation set. During training, we apply common data preprocessing strategies of random cropping and flipping. The results in Table VIII illustrate the significant performance improvement induced by self-attention blocks when introduced into self-proliferation architectures. SP&A-Net has a top-5 error of 6.25% which is superior to both its direct counterpart ResNet with attention block. In terms of model efficiency, SP&A-Net also have 8% fewer floating-point operations (FLOPS) and 20% fewer parameters.

TABLE VIII
COMPARISON OF STATE-OF-THE-ART METHODS ON IMAGENET

| Model | Top-1 err. | Top-5 err. | #Params (M) | FLOPs (G) |
|---|---|---|---|---|
| ResNet50+SE Block | 23.31% | 6.63% | 25.82 | 38.69 |
| ResNet50+NL Block | 24.17% | 7.71% | 39.84 | 59.61 |
| ResNet50+SNL Block | 24.33% | 7.56% | 36.15 | 39.32 |
| ResNet50+GC Block | 24.28% | 7.75% | 34.32 | 39.31 |
| SPNet (without attention) | 23.12% | 6.32% | 20.48 | 34.79 |
| SP&A-Net (with self-attention) | 23.06% | 6.25% | 20.62 | 35.59 |

## V. CONCLUSION

In this paper, we explored the effectiveness of image classification models that are based on self-attention and self-proliferation. Our framework achieves competitive predictive performance on defect pattern classification tasks while requiring fewer parameters and floating-point operations than the corresponding convolution baselines. Experimental results yielded many significant findings. First, the major finding is that self-proliferation mechanism is particularly powerful and substantially outperforms traditional convolutional neural networks. Secondly, our networks based on self-attention match or outperform convolutional baselines. This indicates that the success of convolution in computer vision is not inextricably tied to the long-range dependencies. By attention mechanism, we can develop better methods for capturing geometries. Finally, additional work on proposing new attention forms that can capture low level features can make attention effective in the early layers of networks. This suggests that self-attention, which generalizes convolution, may yield strong accuracy gains across applications in defect pattern classification.

## REFERENCES

[1] V. Borisov and J. Scheible, "Lithography Hotspots Detection Using Deep Learning," 2018 15th International Conference on Synthesis Modeling Analysis and Simulation Methods and Applications to Circuit Design (SMACD), Aug. 2018.

[2] T. Nakazawa and D. V. Kulkarni, "Wafer Map Defect Pattern Classification and Image Retrieval Using Convolutional Neural Network," IEEE Trans. Semicond. Manuf., vol. 31, no. 2, pp. 309-314, Jan. 2018.

[3] Y. F. Yang and M. Sun, "Double Feature Extraction Method for Wafer Map Classification Based on Convolution Neural Network," 2020 31st Annual SEMI Advanced Semiconductor Manufacturing Conference (ASMC), pp. 1-6, Aug. 2020.

[4] K. Simonyan and A. Zisserman, "Very deep convolutional networks for large-scale image recognition," arXiv preprint arXiv:1409.1556, 2014.

[5] S Ioffe, C Szegedy, et al, "Batch Normalization: Accelerating Deep Network Training by Reducing Internal Covariate Shift," Proceedings of the 32nd International Conference on International Conference on Machine Learning, Vol.37, pp. 448–456, 2015.

[6] K. He, X. Zhang, S. Ren and J. Sun, "Deep Residual Learning for Image Recognition," 2016 IEEE Conference on Computer Vision and Pattern Recognition (CVPR), pp. 770-778, 2016.

[7] X. Wang, R. Girshick, A. Gupta and K. He, "Non-local Neural Networks," 2018 IEEE/CVF Conference on Computer Vision and Pattern Recognition, pp. 7794-7803, 2018.

[8] H. Hu, Z. Zhang, Z. Xie and S. Lin, "Local Relation Networks for Image Recognition," 2019 IEEE/CVF International Conference on Computer Vision (ICCV), pp. 3463-3472, 2019.

[9] A. G. Howard, M. Zhu, B. Chen, D. Kalenichenko, W. Wang, T. Weyand, et al., "MobileNets: Efficient Convolutional Neural Networks for Mobile Vision Applications," arXiv preprint arXiv:1704.04861, 2017.

[10] M. Sandler, A. Howard, M. Zhu, A. Zhmoginov and L. Chen, "MobileNetV2: Inverted Residuals and Linear Bottlenecks," 2018 IEEE/CVF Computer Vision and Pattern Recognition Conference (CVPR), pp. 4510-4520, Jun. 2018.

[11] A. Howard et al., "Searching for MobileNetV3," 2019 IEEE/CVF International Conference on Computer Vision (ICCV), pp. 1314-1324, Nov. 2019.

[12] J. Hu, L. Shen and G. Sun, "Squeeze-and-Excitation Networks," 2018 IEEE/CVF Conference on Computer Vision and Pattern Recognition (CVPR), pp. 7132-7141, Jun. 2018.

[13] K. Han, Y. Wang, Q. Tian, J. Guo, C. Xu and C. Xu, "GhostNet: More Features From Cheap Operations," 2020 IEEE/CVF Conference on Computer Vision and Pattern Recognition (CVPR), pp. 1577-1586, 2020.

[14] A. Vaswani, N. Shazeer, N. Parmar, J. Uszkoreit, L. Jones, A. N. Gomez, Ł. Kaiser, and I. Polosukhin, "Attention is all you need," Neural Information Processing Systems (NeurIPS), pp. 5998–6008, 2017.

[15] J. Devlin, M. Chang, K. Lee, and K. Toutanova, "BERT: pre-training of deep bidirectional transformers for language understanding," North American Chapter of the Association for Computational Linguistics (NAACL), pp. 4171–4186, 2019.

[16] L. Chen et al., "SCA-CNN: Spatial and Channel-Wise Attention in Convolutional Networks for Image Captioning," 2017 IEEE Conference on Computer Vision and Pattern Recognition (CVPR), pp. 6298-6306, 2017.

[17] J-J. Liu, Q. Hou, M.-M. Cheng, C. Wang, J. Feng, "Improving Convolutional Networks With Self-Calibrated Convolutions," 2020 IEEE/CVF Conference on Computer Vision and Pattern Recognition (CVPR), pp. 10093-10102, 2020.

[18] Y. Cao, J. Xu, S. Lin, F. Wei and H. Hu, "GCNet: Non-Local Networks Meet Squeeze-Excitation Networks and Beyond," 2019 IEEE/CVF International Conference on Computer Vision Workshop (ICCVW), pp. 1971-1980, 2019.

[19] Y. Li, X. Jin, J. Mei, X. Lian, L. Yang, C. Xie, Q. Yu, Y. Zhou, S. Bai, L. Yuille, "Neural Architecture Search for Lightweight Non-Local

Networks," 2020 IEEE/CVF Conference on Computer Vision and Pattern Recognition (CVPR), pp. 10294-10303, 2020.

[20] K. He, X. Zhang, S. Ren and J. Sun, "Deep Residual Learning for Image Recognition," 2016 IEEE Conference on Computer Vision and Pattern Recognition (CVPR), pp. 770-778, Jun. 2016.

[21] X. Glorot and Y. Bengio, "Understanding the difficulty of training deep feedforward neural networks," In Proceedings of the thirteenth international conference on artificial intelligence and statistics, pages 249--256, 2010.

[22] Y. Sun et al., "Circle Loss: A Unified Perspective of Pair Similarity Optimization," 2020 IEEE/CVF Conference on Computer Vision and Pattern Recognition (CVPR), pp. 6397-6406, Jun. 2020.

## GITHUB REPOSITORY

In this Repository (*https://github.com/Yfvangd/SPA*), the code related to this paper is placed. Due to the principle of data confidentiality protection, we did not include the defect pattern image dataset. Instead, we used the public data material CIFAR-10 as a model demonstration.

· *Self_Proliferate.py* is used to generate more feature maps (as section 3-A).

· *Self_Attention.py* is used to capturing the long-range dependencies of the feature map (as section 3-B).

· *Self_Proliferate_and_Attention.py* follow the spirit of MobileNetV2, "capture features in high dimensions and transfer information in low dimensions", to make the network more efficient. (as section 3-C).

· *SPA_Net.py* is the overall network architecture of SP&A-Net (as section 3-D).

· *CircleLoss.py* is used to estimate the loss rate during model training with two elemental deep feature learning approaches: class-level labels and pair-wise labels (as section 3-E).

· *SP&A-Net-Test-Run.ipynb* is in the form of a Jupyter Notebook as a simple display with CIFAR-10 as the training object.